\documentclass{article}

\usepackage{PRIMEarxiv}

\usepackage[utf8]{inputenc} 
\usepackage[T1]{fontenc}    
\usepackage{hyperref}       
\usepackage{url}            
\usepackage{booktabs}       
\usepackage{amsfonts}       
\usepackage{nicefrac}       
\usepackage{microtype}      
\usepackage{lipsum}
\usepackage{fancyhdr}       
\usepackage{graphicx}       
\graphicspath{{media/}}     
\usepackage{algorithm}
\usepackage{algorithmic}
\usepackage{graphicx}
\usepackage{tcolorbox}
\usepackage{xcolor}      
\usepackage{booktabs} 

\title{KeyKnowledgeRAG ($K^2RAG$): An Enhanced RAG method for improved LLM question-answering capabilities
\thanks{\textit{\underline{Citation}}: 
\textbf{Authors. Title. Pages.... DOI:000000/11111.}} 
}

\author{
  Hruday Markondapatnaikuni \\
  University of Sydney \\
  \texttt{smar0241@uni.sydney.edu.au} \\
   \And
  Basem Suleiman \\
  University of New South Wales \\
  \texttt{b.suleiman@unsw.edu.au} \\
   \And
    Abdelkarim Erradi \\
  Qatar University \\
  \texttt{erradi@qu.edu.qa} \\
   \And
    Shijing Chen \\
  University of New South Wales \\
  \texttt{arthur.chen@unsw.edu.au} \\
}

\begin{document}
\maketitle

\begin{abstract}
Fine-tuning is an immensely resource expensive process when trying to retrain Large Language Models (LLMs) to have access to a larger bank of knowledge. To alleviate this issue there have been many different fine-tuning techniques proposed which have shown good progress in trying to reduce time and computational resources to achieve fine-tuning but with LLMs becoming more intelligent and larger, this issue continues to arise. Hence a new method of enabling knowledge expansion on LLMs had to be devised. Retrieval-Augment-Generate (RAG) is a class of techniques where information is stored in a database and appropriate chunks of information are retrieved to help answer the question. However there are many limitations to naive RAG implementations. This paper proposes the KeyKnowledgeRAG ($K^2RAG$) framework to address the scalability and answer accuracy limitations associated with naive RAG implementations. This framework takes inspiration from divide-and-conquer ideology, and combines dense and sparse vector search, knowledge graphs and text summarization to help address these limitations. The framework also involves a data pre-processing step to reduce training times. The MultiHopRAG dataset was used for evaluation where the our implemented $K^2RAG$ pipeline was trained with the document corpus and then evaluated with the test data. Results have shown that there is an improvement in answering questions compared to common Naive RAG implementations where $K^2RAG$ achieved the highest mean answer similarity at \textbf{0.57} and also was able to answer more questions with more ground truth similarity with a highest $Q_3$ quartile at \textbf{0.82}. Additionally our proposed framework is trainable much faster due to the inclusion of a train data corpus summarization step which reduces training times of the individual components by \textbf{93\%} on average. Furthermore $K^2RAG$ has shown to operate even faster than a traditional knowledge graph based naive RAG implementation with a mean execution times reduced by up to \textbf{40\%}. In addition to superior question-answering capabilities, $K^2RAG$ offers being a more scalable solution with VRAM requirements reduced by \textbf{3 times} in comparison to implementations of several naive RAG pipelines evaluated in this paper. Hence $K^2RAG$ can help companies in sophisticated decision-making through implementing a more lightweight and robust question-answering systems built on internal documents.
\end{abstract}


\section{Introduction}
\textbf{Motivation:}
Customizing LLMs to be more knowledgeable on specific areas has been a popular task to accomplish in recent years. A technique known as fine-tuning has been proposed and popularized through various different methods of fine-tuning coming up over the past few years such as FLAN \cite{wei2022finetunedlanguagemodelszeroshot}, LoRA \cite{hu2021loralowrankadaptationlarge} and QLoRA \cite{dettmers2023qloraefficientfinetuningquantized} which all aim to address different setbacks associated with the LLM fine-tuning process such as improving the final efficacy as well as reducing computational resources. 

 Fine-tuning LLMs is usually done on full LLMs and is generally not advised to fine-tune on or convert pre-finetuned models into quantized models due to potential loss of knowledge retention and accuracy which can be attributed to lower precision of the model's weights. Even though existing techniques like QLoRA have shown results which indicate minimal performance degradation upon the resulting fine-tuned quantized models which are based on their Guanaco models, this is not applicable to all LLMs \cite{dettmers2023qloraefficientfinetuningquantized}. For example an empirical study \cite{huang2024empiricalstudyllama3quantization} evaluated different LLAMA3 quantizations and fine-tune quantizations has shown a significant reduction in LLAMA3-8B scores on the MMLU benchmarks when compared to the standard full 16-bit model and a QLoRA fine-tune-quantization to 4bits. This is because although all LLMs follow the basic transformer architecture there are still differences in model designs which can impact the quality of fine-tunes and hence even previously considered efficient methods such as QLoRA face their limitations with newer models \cite{huang2024empiricalstudyllama3quantization}.

Hence a major setback from all fine-tuning methods is that they are still extremely time and computational resource intensive in order to yield satisfactory results. Hence a new approach needed to be devised whereby LLMs could gain access to more knowledge  with more computational efficiency.

\begin{figure}[ht] \centering
\includegraphics[scale=0.35]{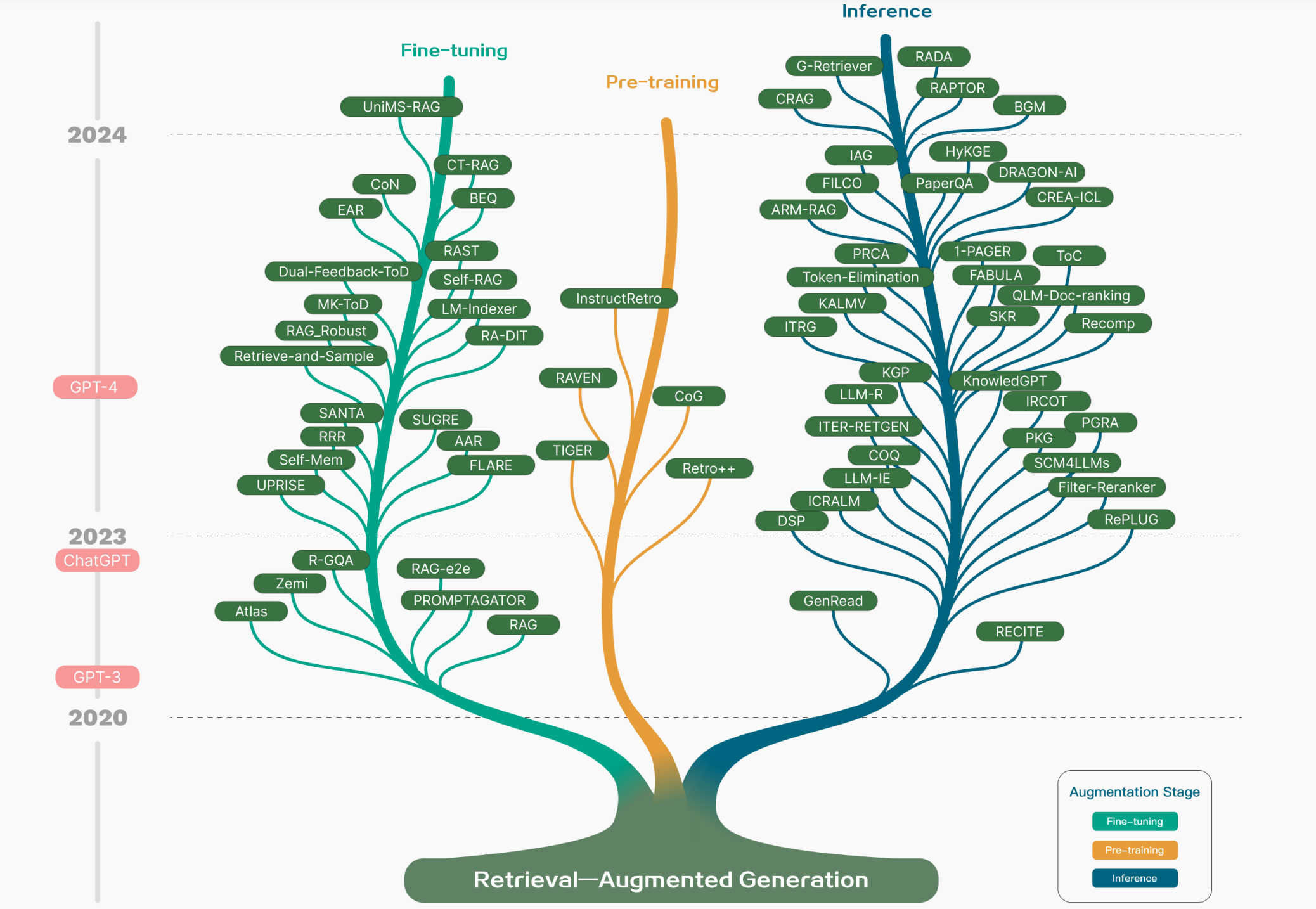}
\caption{Evolution of RAG since 2020 \cite{gao2024retrievalaugmentedgenerationlargelanguage}.}
\label{fig:rag_evo}
\end{figure}

A class of techniques known as Retrieval Augment Generate (RAG)  systems have been devised as an alternative to fine-tuning where they do not employ any step where a base LLM is modified as in the fine-tuning method but rather knowledge is extracted from documents and indexed in a knowledge holding system to then retrieve \cite{lewis2021retrievalaugmentedgenerationknowledgeintensivenlp}. This helps eliminate the computationally expensive associated with fine-tuning methods and also comes with the added benefit of having a dynamic knowledge base which could be updated quicker than fine-tuning methods.

Since 2020 the concept of RAG has since evolved to include many different implementations and approaches to performing information retrieval to enhance LLM answering capabilities on new datasets [Figure \ref{fig:rag_evo}]. This highlights how the benefits gained from RAG are being researched in depth to further optimize this process.

\begin{figure}[ht] \centering
\includegraphics[scale=0.75]{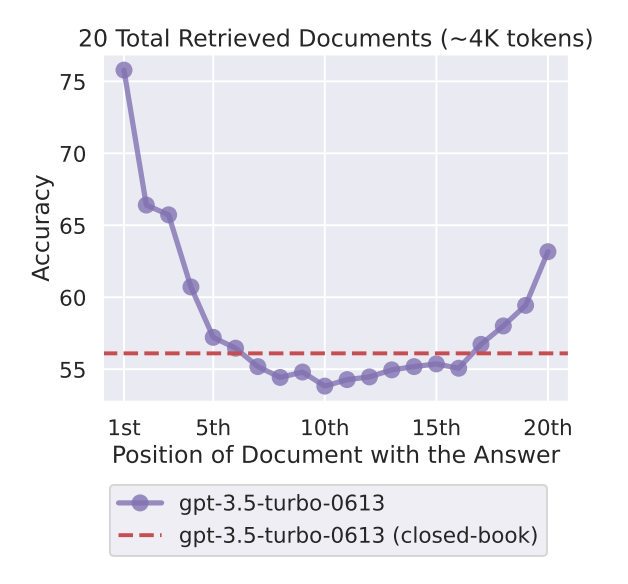}
\caption{Graph showing how LLMs prefer extracting information from top or bottom of context to answer a question \cite{liu2023lostmiddlelanguagemodels}.}
\label{fig:needle-in-haystack-bias}
\end{figure}

\textbf{Answer Accuracy:} Naive RAG implementations often struggle with the "Needle-in-a-Haystack" problem \cite{laban2024summaryhaystackchallengelongcontext}, where the generator model fails to provide accurate answers due to an inability to extract relevant information from irrelevant or excessively long contexts. This issue typically arises from poorly optimized chunk sizes and retrieval processes \cite{wang2024searchingbestpracticesretrievalaugmented}. Optimizing chunk size is critical and challenging, as smaller chunks risk losing essential context, while larger chunks may introduce irrelevant information and unnecessarily increase context length [Figure \ref{fig:needle-in-haystack-bias}].
Optimizing the retrieval process to maximize the relevance of retrieved information blocks is crucial. This ensures that irrelevant data, which could act as noise and obscure the correct information, does not compromise the generator LLM's ability to accurately answer the question. Another limitation of naive RAG systems is that embeddings which are similar but not actually relevant to answer the question might be retrieved \cite{rossirelevancefiltering} as the original sentences might be semantically similar or carry a similar meaning but might not be right in helping answer the question in that context. 

\textbf{Scalability:} Naive implementations of Retrieval-Augmented Generation (RAG) often rely on 16-bit floating-point large language models (LLMs) for the generation component. However, this approach introduces significant scalability challenges due to the increased memory demands required to host the LLM as well as longer inference times due to using a higher precision number type. To enable more efficient scaling, it is crucial to integrate methods or techniques that reduce the memory footprint and inference times of generator models. Quantized models offer more scalable solutions due to less computational requirements, hence when developing RAG systems we should aim to use quantized LLMs for more cost effective deployment as compared to a full fine-tuned LLM whose performance might be good but is more expensive to deploy due to higher memory requirements. A quantized LLM's role in the RAG pipeline itself should be minimal and for means of rewriting retrieved information into a presentable fashion for the end users to interpret.

Additionally, in naive RAG implementations, information store components, such as Knowledge Graphs, face lengthy training times for their creation and updating. This creates scalability challenges, as it increases the downtime required to update the information store with new knowledge in a production environment. Hence an effective data pre-processing step should be introduced to reduce the training corpus size and hence training times while preserving as much information as possible.

\noindent\textbf{Solution:} We propose the $K^2RAG$ framework which addresses the following 4 research goals based on the aforementioned answer accuracy and scalability issues characteristic of naive RAG implementations:

\label{goal1:reducetraintimes}\noindent\textbf{Goal 1.} Reduce information store such as Knowledge Graph and Spare and Dense vector database creation times. 

\label{goal2:reduceneedleinhaystack}\noindent\textbf{Goal 2.} Reduce chances of LLMs suffering from "Needle in Haystack" Problem. 

\label{goal3:increaseretrieveaccuracy}\noindent\textbf{Goal 3.} Increase rate of retrieving relevant passages for answering the question. 

\label{goal4:reduceresources}\noindent\textbf{Goal 4.} Alleviate time and computational cost associated with full LLMs.

$K^2RAG$ incorporates various different concepts covered in the literature in a novel way to improve the performance and offer a more resource efficient and accurate RAG pipeline compared to naive RAG methods outlined in the literature.

\section{Related Works}
Several studies have explored and implemented advanced RAG (Retrieval-Augmented Generation) systems to overcome the limitations of the naive RAG approach discussed in the introduction. A significant focus of these works lies in developing novel methods to enhance vector database-centric retrieval systems, evaluated across a diverse set of metrics pertinent to vector-based retrieval. Additionally, some studies assess the performance of RAG solutions leveraging Knowledge Graphs \cite{Hogan_2021}, while one investigation examines the outcomes of naively combining Knowledge Graph-based and semantic search results for question-answering tasks. These related works also highlight certain limitations, ranging from increased resource demands to the need for novel strategies or enhancements to improve system performance.

Addressing the limitations of naive Retrieval-Augmented Generation (RAG) pipelines has led to the integration of hybrid retrieval methods that combine sparse (keyword-based) and dense (semantic) retrievers.

\textbf{Spare and Dense Retriever RAG methods: } Hybrid approaches, such as the "Blended RAG" framework proposed by Sawarkar et al. \cite{sawarkar2024blendedragimprovingrag}, demonstrate that combining multiple retrievers can improve the retrieval of relevant passages for clearly defined queries. This framework integrates six distinct retrievers—including traditional sparse methods (e.g., BM25) and dense semantic search techniques—selected based on their performance across benchmark datasets like Natural Questions (NQ) \cite{kwiatkowski-etal-2019-natural}, TREC-COVID \cite{wang-etal-2020-cord}, SqUAD \cite{rajpurkar2016squad100000questionsmachine}, and HotPotQA \cite{yang2018hotpotqadatasetdiverseexplainable}.

A key strength of this hybrid search technique is its ability to return relevant passages, aligning well with Research Goal 3 for well-specified queries. However, these methods do not fully address Research Goal 3 when handling vaguer questions. In such cases, the hybrid approach can retrieve extraneous or unnecessary information, thus failing to eliminate the "needle in a haystack" problem. This shortcoming is evident in evaluations on the CoQA dataset \cite{reddy-etal-2019-coqa}, where retrieval accuracies ranged only from 45\% to 49\%. The reliance on document metadata during retrieval—a common feature in these systems—further complicates matters, as such metadata is often minimal in real-world scenarios (e.g., limited to document names and creation dates), thereby increasing both the creation time of the information store (contradicting Research Research Goal 1) and the risk of returning irrelevant passages (contradicting Research Research Goal 3) leading to longer contexts (contradicting Research Goal 2).

\begin{figure}[ht] \centering
\includegraphics[scale=0.3]{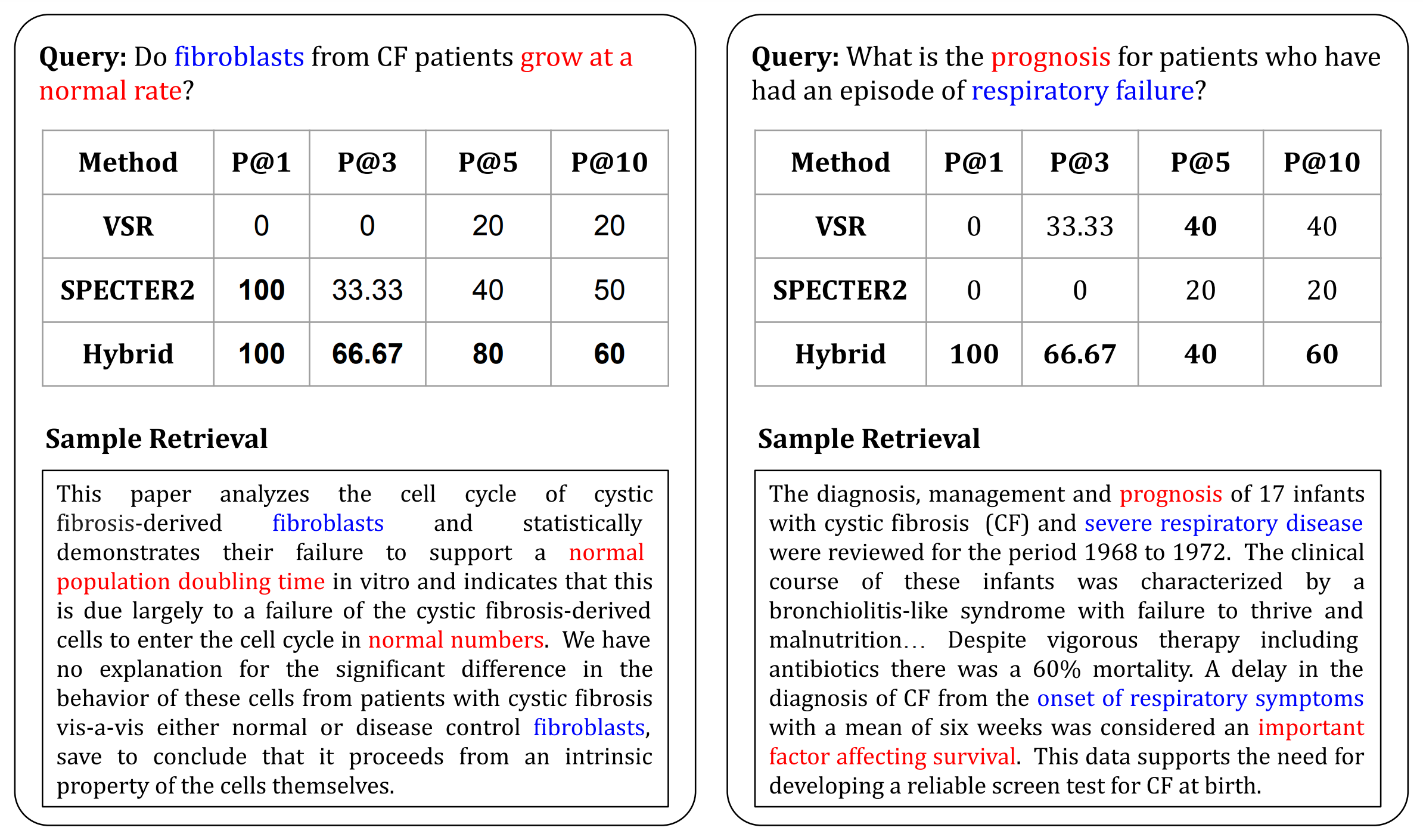}
\caption{Results of Priyanka Mandikal et al. on 2 example queries \cite{mandikal2024sparsemeetsdensehybrid}.}
\label{fig:hybrid_search_mandika_results}
\end{figure}

In contrast, recent work by Mandikal et al.  \cite{mandikal2024sparsemeetsdensehybrid} has advanced a hybrid retrieval model that integrates a sparse TF-IDF retriever with a dense retriever based on fine-tuned SPECTER2 embeddings \cite{singh2023scirepevalmultiformatbenchmarkscientific}. Their approach employs a weighted scoring mechanism, controlled by a hyperparameter $\lambda$, to balance the contributions of the two retrieval methods. Empirical evaluations on a medical database concerning Cystic Fibrosis indicate that while dense retrieval alone underperforms, incorporating approximately 20\% of the sparse score (i.e., setting $\lambda = 0.2$ for the sparse component) significantly improves document retrieval accuracies across different retrieval ranks [Figure \ref{fig:hybrid_search_mandika_results}]. Crucially, this method indexes documents solely based on their content, thereby avoiding the pitfalls of metadata dependency. This contributes to reducing information store creation times (Research Goal 1) and partially mitigates the retrieval of unnecessary information (Research Goal 3), although it does not entirely resolve the challenge of handling vaguer queries.

In summary, while hybrid retrieval approaches represent a significant advancement by returning more relevant passages for well-defined queries complying with Research Goal 3. However they remain imperfect in filtering out unnecessary information for vaguer questions which can lead to longer contexts. This limitation underscores the need for further refinement to minimize the "needle in a haystack" problem in large-scale, real-world applications and make a pipeline which is compliant with Research Goal 2 as well.

\textbf{Reranker RAG methods: }To address the limitation which Hybrid-Search methods have in adhering to Research Goal 2 for vaguer questions, Reranker methods have been developed to reduce the size and improve the quality of the final context.

\begin{figure}[ht] \centering
\includegraphics[scale=0.15]{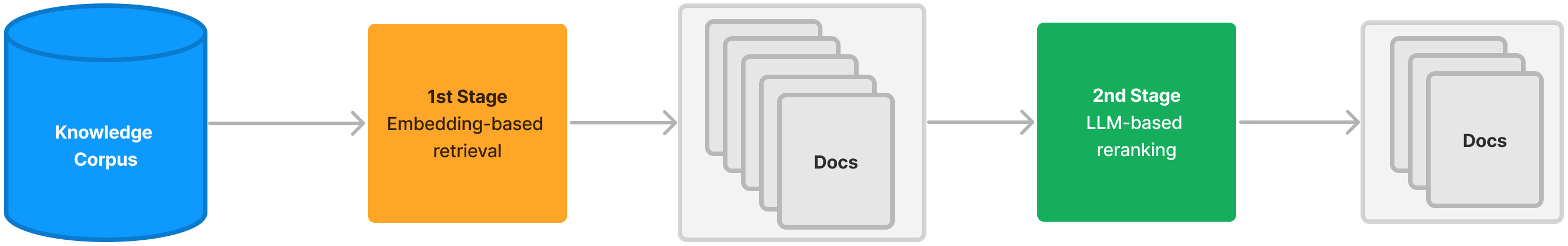}
\caption{A high level view of what a RAG pipeline with a reranker looks like.}
\label{fig:reranker_rag_high_level_view}
\end{figure}

Usually this is a process called 2-step retrieval where the first step involves retrieving documents such as through sparse or dense retriever techniques and then a second stage where an LLM based reranker comes into play to decide which of the documents retrieved are most relevant to the query and sorts them in order [Figure \ref{fig:reranker_rag_high_level_view}] before passing it to an LLM for generation. 

Prior to the formal conceptualization of Retrieval-Augmented Generation (RAG), information retrieval research explored the use of document rerankers. For example, Noguiera et al.'s 2020 study, Document Ranking with a Pretrained Sequence-to-Sequence Model \cite{nogueira-etal-2020-document}, proposed employing a sequence-to-sequence model as a document reranker. This work evaluated multiple fine-tuned T5 models of varying sizes, as well as a fine-tuned BERT model, on documents initially retrieved using the sparse BM25 keyword method. The incorporation of these rerankers led to improvements in retrieval performance across the Robust04, Core17, and Core18 datasets. However, a critical limitation of this framework was its reliance on large models, leaving unresolved whether the observed performance gains were due to the reranking strategy or simply the use of more powerful models.

Subsequent advancements in RAG have integrated BERT-based rerankers into new pipelines. Michael Glass et al.'s Re\textsuperscript{2}G framework \cite{glass2022re2gretrievererankgenerate} embeds a BERT-based reranker—similar to that of Noguiera et al.—within a RAG system. This integration has shown notable improvements in metrics such as Recall (for 5 documents), overall recall, precision, and accuracy on several datasets, as indicated by performance on the KILT leaderboard. Nevertheless, a significant drawback of this approach is the necessity to host an additional model, which directly conflicts with Goal 4: alleviating the monetary and computational costs associated with deploying full-scale LLMs.

Further refinement in reranking approaches is demonstrated by Xuegang Ma et al., who fine-tuned a LLaMA-2-7B model for reranking in their RankLLaMA system \cite{ma2023finetuningllamamultistagetext}. Although RankLLaMA achieves superior reranking performance, it again relies on a large model comparable to LLaMA-2-7B. Such reliance on resource-intensive models undermines the overarching objective of Goal 4, as the additional computational and financial burden associated with hosting these rerankers contradicts the need for efficiency and cost reduction.

In summary, while reranker models can mitigate the “needle in a haystack” problem (Goal 2) by filtering out least relevant information, they are not a sustainable solution. Their requirement for additional large-scale models significantly increases computational and monetary costs, thereby violating Goal 4. Consequently, alternative strategies that achieve comparable retrieval performance without incurring such overhead are essential for the advancement of practical and efficient RAG systems.

\textbf{Knowledge Graph RAG methods: } Knowledge graph–based solutions have been proposed as a promising alternative for enhancing the RAG process. The core idea behind employing knowledge graphs is to address semantic search challenges by organizing data according to intrinsic relationships. For example, although not originally applied within a RAG context, Dawei Cheng et al.'s 2020 work on constructing knowledge graphs for a financial price prediction tool \cite{CHENG2022108218} illustrates how grouping semantically related data into distinct sub-networks enables a query—such as “who quit Apple”—to retrieve only the pertinent subset of information, rather than returning all semantically similar sentences (e.g., both “Steve Jobs quits Apple” and “David Peter leaves Starbucks”). This targeted retrieval helps mitigate the "needle in a haystack" problem (Goal 2) and increases the rate of retrieving relevant passages (Goal 3).

Building on these insights, recent research has directly integrated knowledge graphs within RAG frameworks. For instance, Zhentao Xu et al. \cite{Xu_2024} developed a system for LinkedIn’s customer question answering service by first creating a custom knowledge graph and then retrieving related nodes. The retrieved information is subsequently refined using GPT-4 to produce coherent responses. When compared with a baseline semantic search system, this pipeline significantly improved retrieval metrics—including Mean Reciprocal Rank, Recall, and Normalized Discounted Cumulative Gain—as well as evaluation measures such as BLEU, METEOR, and ROUGE, thereby supporting Goals 2 and 3.

However, despite these promising advances, knowledge graph–based approaches fall short of meeting Goal 1. The process of constructing and training knowledge graphs is computationally intensive and time-consuming, which directly conflicts with the objective of reducing information store creation times. 

\begin{figure}[ht] \centering
\includegraphics[scale=0.3]{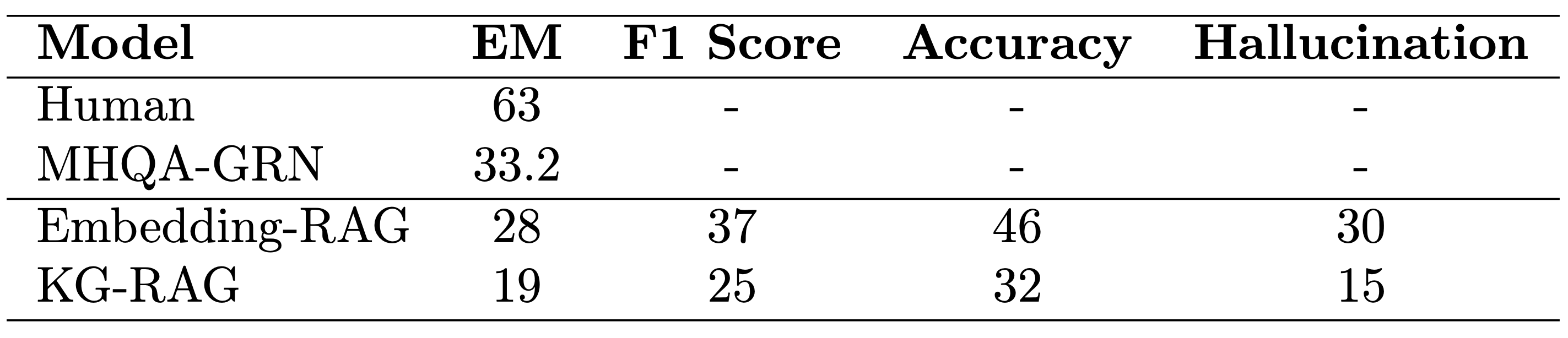}
\caption{Results comparing knowledge graph framework with semantic search methods \cite{sanmartin2024kgragbridginggapknowledge}.}
\label{fig:kg-rag-results}
\end{figure}

For example, Diego Sanmartín’s 2024 work \cite{sanmartin2024kgragbridginggapknowledge} demonstrated a Knowledge Graph RAG system that leverages LLMs to extract entity-relationship triples from unstructured text and employs a “Chain of Explorations” retrieval strategy. While this approach shows potential for handling vague queries, the high training times required to build and maintain the knowledge graph significantly increase the overall overhead. Moreover, the compressed storage format—using entity-relationship triples—can lead to a loss of detailed contextual information, thereby limiting the system’s ability to fully expand upon relevant topics during retrieval hence resulting in even lower EM, F1, Accuracy scores with only an improvement in Hallucination compared to a standard Semantic Search method [Figure \ref{fig:kg-rag-results}].

In summary, while knowledge graph–based solutions offer clear benefits in terms of improving semantic search and enhancing the retrieval of relevant passages (addressing Goals 2 and 3), they do not currently satisfy Goal 1 due to their high training times and associated computational costs. Addressing these inefficiencies is essential for realizing the full potential of knowledge graphs within efficient and practical RAG frameworks.

\textbf{Combining Knowledge Graphs and Vector based Search methods: } The limitations inherent in a purely knowledge graph–based RAG framework suggest that combining the strengths of knowledge graphs with traditional semantic search methods may yield superior results. Bhaskarjit Sarmah et al. \cite{sarmah2024hybridragintegratingknowledgegraphs} propose such an integrated approach through their HybridRAG framework. This framework concatenates outputs from a knowledge graph query—implemented using Microsoft’s GraphRAG \cite{edge2024localglobalgraphrag} with content indexed by extracting entity relationships in chunk sizes of 1024—with results from a semantic search, which similarly employs chunk sizes of 1024 and the text-embedding-ada-002 model. The aggregated context is then provided to GPT-3.5 to generate the final answer. The dataset for this study was constructed by web scraping earnings call documents of companies listed on the NIFTY 50, and evaluation was performed using the RAGAS evaluation platform \cite{es2023ragasautomatedevaluationretrieval}.

The HybridRAG framework demonstrated improvements in metrics such as Faithfulness, Answer Relevancy Score, and Context Recall. These enhancements support Goal 2 by reducing the "needle in a haystack" problem and Goal 3 by increasing the rate of retrieving relevant passages for answering questions. However, the framework underperformed in Context Precision when compared to standalone implementations of both knowledge graph–based and vector database–based RAG systems. This shortcoming indicates that the current integration does not fully leverage the advantages of traditional vector search techniques, particularly hybrid search techniques, which is critical for fully meeting Goal 3.

\section{Methodology: }
\subsection{$K^2RAG$ Framework}
To meet the research goals, the $K^2RAG$ framework implements four techniques:

\begin{itemize}
    \item \textbf{1. Knowledge Graph:} $K^2RAG$ employs a knowledge graph component to organize and interconnect topics within the corpus [Figure \ref{fig:$K^2RAG$outline}(a)]. Unlike traditional chunking approaches, which struggle to balance the trade-off between providing sufficient context and avoiding irrelevant information, the knowledge graph addresses this issue by structuring content into interconnected nodes which enables knowledge-rich and more context-aware retrieval capable of handling complex or vague queries that vector-based search methods often struggle with.
    
    \item \textbf{2. Hybrid Search:} To minimize false positive embeddings—where semantically similar but irrelevant chunks are retrieved—$K^2RAG$ integrates a hybrid retriever [Figure \ref{fig:$K^2RAG$outline}(b)]. Inspired by Mandikal et al.'s paper \cite{mandikal2024sparsemeetsdensehybrid}, we weight dense vector and sparse vector retrieval at an optimized 80\%/20\% ratio. This hybrid method significantly improves retrieval accuracy, helping reduce noise by ensuring the most relevant content is retrieved.
    
    \item \textbf{3. Summarization:} To combat the "Needle-in-Haystack" issue, $K^2RAG$ incorporates summarization at multiple stages in the pipeline [Figure \ref{fig:$K^2RAG$outline} (B) and (E)]. To speed up indexing, documents are summarized before being indexed into vector and knowledge graph stores. Additionally retrieved content is summarized at query time to further refine the context provided to the LLM. By summarizing at both indexing and retrieval, we ensure concise, high-quality input for generation. 
    
    \item \textbf{4. Lightweight Models:} For VRAM resources efficiency, $K^2RAG$ leverages a lightweight Longformer-based text summarizer \cite{guo2022longt5efficienttexttotexttransformer}, along with quantized LLMs to produce a low VRAM and resource-efficient pipeline without compromising output quality.
\end{itemize}

\subsubsection{Corpus Summarization Process }
\begin{figure}[ht] \centering
\includegraphics[scale=0.7]{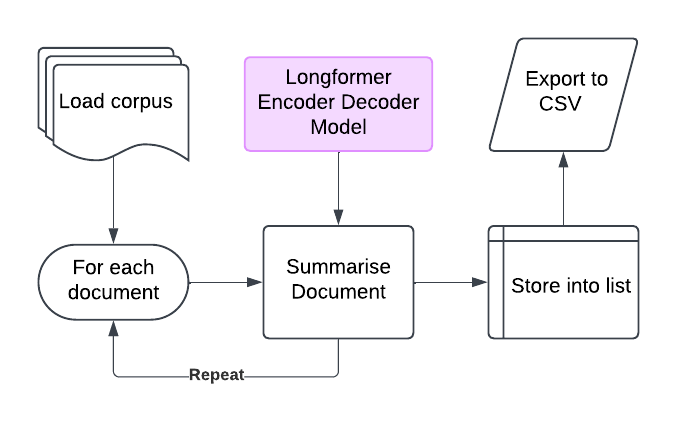}
\caption{How to generate summaries of corpus.}
\label{fig:summarisationprocessgraph}
\end{figure}

To address research goal 1, we decrease training times of the $K^2RAG$ framework compared to naive RAG frameworks by summarizing the training corpus to reduce its size while preserving as much information as possible. Hence a Longformer based model \cite{guo2022longt5efficienttexttotexttransformer} fine tuned for long text summarization is used as it's self-attention mechanism is highly optimized for capturing and processing large chunks of texts compared to those seen in traditional transformer models. This helps speed up training times of information store components used in $K^2RAG$. To create our summarized corpus we iterate over each article in the corpus and then summarize it using the loaded in Longformer summarizer model. Then these results are stored into a CSV format from which then they can be loaded in for training the RAG pipeline [Algorithm \ref{algo:summarised_data}] [Figure \ref{fig:summarisationprocessgraph}].

\begin{algorithm}
\caption{Generating Summarised Corpus}
\begin{algorithmic}[1]
\STATE Load in $summariser$ $model$
\STATE Load in $corpus$
\STATE Initialise $summarised$ $corpus$ as empty list
\FOR{$document$ in $corpus$}
    \STATE Generate $summary$ of $document$ using $summarised$ $model$
    \STATE Append $summary$ to $summarised$ $corpus$ list
\ENDFOR
\end{algorithmic}
\label{algo:summarised_data}
\end{algorithm}

\subsubsection{Indexing }
The chunking strategy to generate a set of chunks $C_d$ from a single document or text $d$ is defined as:

\begin{equation}
C_{d} = \bigcup_{i=1}^{L_t}\bigg\{\left\{x_{((i-1)*(S-O))+j} \mid j = 1, 2, \ldots, S \right\}\bigg\},\; L_t = \left\lfloor \frac{Nt_d - O}{S - O} \right\rfloor
\label{formula:chunkdoc}
\end{equation}

where $S$ is the desired chunk size, $O$ is the desired chunk overlap, $Nt_d$ is the total number of tokens in the document $d$. The inner set denotes a chunk constructed from tokens $x_{((i-1)*(S-O))+1}$ to $x_{((i-1)*(S-O))+S}$ and $L_t$ is the limit of tokens to process to ensure the number of chunks of size $S$ and overlap $O$ are maximized.

We then apply the chunking strategy defined in \textbf{(\ref{formula:chunkdoc})} across all documents in the summarized corpus $cor_{sum}$ and take the union of chunks to give us the final set of chunks $C$ to pass to the knowledge graph and vector stores for indexing:

\begin{equation}
C = \bigcup_{d=1}^{|cor_{sum}|} C_d \;|\; d\in cor_{sum}
\label{formula:combineall}
\end{equation}

The configuration used to train all information retrievers is $S=256, O=20$ when chunking for the dense and sparse vector stores indexing, and $S=300, O=100$ when chunking for knowledge graph indexing. 

\begin{figure}[ht] \centering
\includegraphics[scale=0.2]{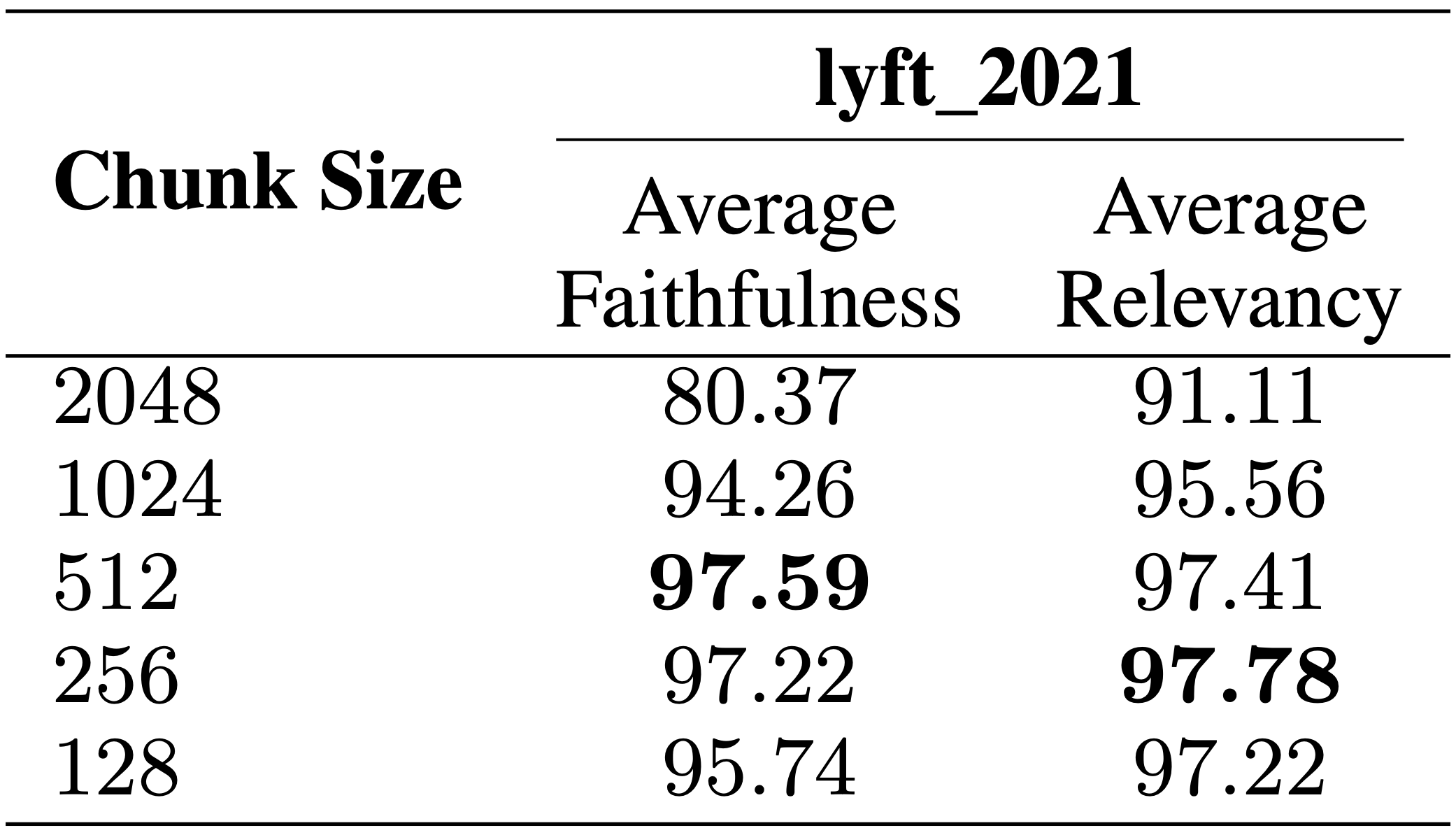}
\caption{Chunk sizes and their effects on Answer Faithfulness and Answer Relevancy.}
\label{fig:best_chunk_sizes}
\end{figure}

This was decided in accordance with Xiaohua Wang et al.'s paper for vector stores indexing which showed 512 and 256 token chunks achieved the best results in Faithfulness and Relevancy metrics \cite{wang2024searchingbestpracticesretrievalaugmented} [Figure \ref{fig:best_chunk_sizes}]. Using the chunk size less than 512 is used to help promote in smaller contexts generated to pass to the generator LLM to address the "Needle-in-Haystack" challenge \cite{laban2024summaryhaystackchallengelongcontext} by ensuring enough information is contained within the chunks retrieved and minimising noisy information.

\subsubsection{Retrieval and Generation }
For all pipelines the following system prompt was used for generation of the final answer:

\begin{tcolorbox}[title=Answer Generation System Prompt, colframe=blue!75!black, colback=blue!10, boxrule=0.5mm, width=\textwidth, rounded corners]
Additional Information:
\textcolor{red}{context from retrieval}

Instruction: You are a smart LLM who gives an answer to the question in as little words as possible using the additional information provided above.

Question: \textcolor{red}{question}

Short Answer:
\label{prompt:answer_gen_prompt}
\end{tcolorbox}

\begin{figure}[ht] \centering
\includegraphics[scale=0.4]{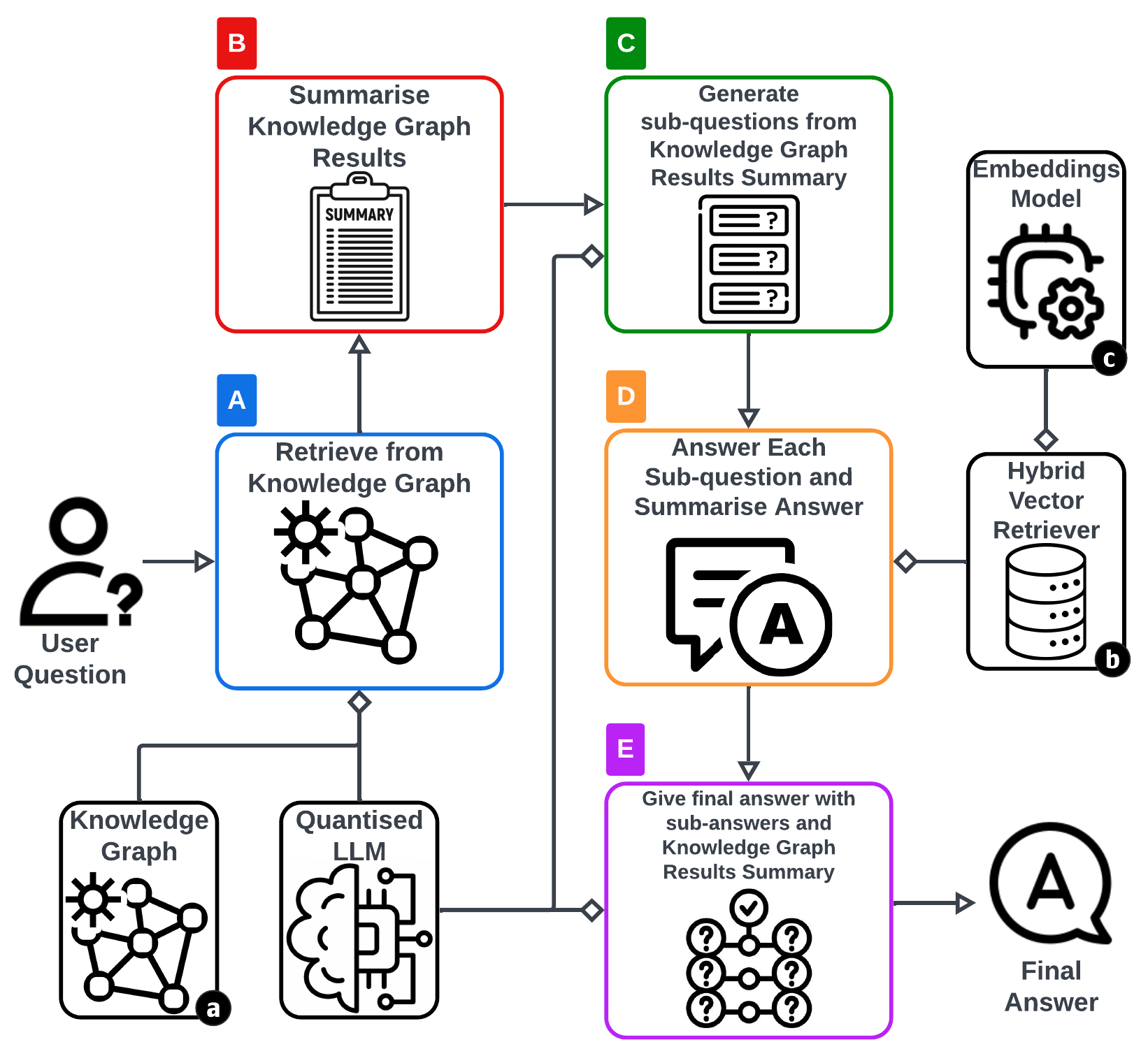}
\caption{$K^2RAG$ framework overview. Component (a): Knowledge Graph. Component (b): Hybrid Retriever containing Dense and Sparse retrievers. Component (c): Embeddings Model. Step (A): Knowledge Graph results. Step (B): Summarise Knowledge Graph results. Step (C): Generating sub-questions from chunks. Step (D): Generating sub-answers for sub-questions. Step (E): Generate the final answer from sub-answers and Knowledge Graph Results Summary.}
\label{fig:$K^2RAG$outline}
\end{figure}

We establish the top $k$ chunks $C_q$ retrieved from the dense and sparse vector databases for a question $q$ is a \textbf{hybrid retriever} [Figure \ref{fig:$K^2RAG$outline} (b)] defined as:

\begin{equation}
C_q = \bigcup_{i=1}^{k} \left\{c \;|\; scr(q, c, \lambda) = \mathrm{max}_i[S] \right\},\;S=\left\{scr(q, c, \lambda) \;|\; c \in C\right\}
\label{formula:hybridretriever}
\end{equation}

where $S$ is the set of scores computed for each chunk in $C$ \textbf{(\ref{formula:combineall})} using a scoring function $scr$ of the question $q$, chunk $c$ and a weighting parameter $\lambda$:

\begin{equation}
scr(q, c, \lambda) = \lambda \frac{e(q) \cdot e(c)}{|e(q)|*|e(c)|} + (1 - \lambda) \frac{z(q) \cdot z(c)}{|z(q)|*|z(c)|}
\label{formula:hybridscore}
\end{equation}

Where $e$ represents the dense embeddings model as a function and $z$ represents the sparse embedding function.

 The first step in the $K^2RAG$ pipeline [Figure \ref{fig:$K^2RAG$outline}] is to use a knowledge graph trained on summarized data to generate an answer containing the identified topics relevant to the query [Figure \ref{fig:$K^2RAG$outline}(A)]. Then a Longformer model is used to summarize the results obtained from the knowledge graph [Figure \ref{fig:$K^2RAG$outline}(B)]. Over the knowledge graph results summary, we split it into chunks of $S=128$ and $O=10$ following our text chunking strategy \textbf{(\ref{formula:chunkdoc})}. This is so that each chunk contains information about a clear topic needed to help answer the question from which for each chunk we use a quantized LLM to create a sub-question out of it [Figure \ref{fig:$K^2RAG$outline}(C)] with the following specialized prompt:

 \begin{tcolorbox}[title=Question Generation System Prompt, colframe=blue!75!black, colback=blue!10, boxrule=0.5mm, width=\textwidth, rounded corners] 
Instruction: Your task is to create a small question out of the below information.

Information: 
\textcolor{red}{chunk from knowledge graph results}

Answer:
\end{tcolorbox}
 
 The sub-question is then used to query the dense and sparse vector databases using the hybrid retriever with $\lambda=0.8$ \cite{mandikal2024sparsemeetsdensehybrid} and $k=10$ \textbf{(\ref{formula:hybridscore})} [Figure \ref{fig:$K^2RAG$outline}(D)] where an embeddings model is used when querying the dense vector database [Figure \ref{fig:$K^2RAG$outline}(c)].

We then use the quantized LLM with the results from the hybrid retriever as context to generate the answer to the sub-question and then store into a list containing all the sub-answers. Once completed for each chunk, we then concatenate all the sub-answers and use the Longformer model to summarize the results so that we can reduce the size of the context [Figure \ref{fig:$K^2RAG$outline}(D)]. Finally, the summarized sub-answers and the summarized knowledge graph results are concatenated and passed as context to the quantized LLM to generate the final answer to the question [Figure \ref{fig:$K^2RAG$outline}(E)].

\begin{algorithm}
\caption{$K^2RAG$ Retrieval and Generation}
\begin{algorithmic}[1]
\STATE Load in $Quantized LLM$
\STATE Define the $generation$ $prompt$ for generation
\STATE Define the $question$ $prompt$ for question creation

\STATE Load in $embeddings model$
\STATE Load in $summariser$ $model$

\STATE Load in $knowledge$ $graph$
\STATE Load in $BM25$ $store$
\STATE Load in $semantic$ $vector$ $store$

\STATE Obtain \textbf{user input}

\STATE Retrieve $knowledge$ $graph$ $output$ from $knowledge$ $graph$ using \textbf{user input} and $embeddings model$ and $Quantized LLM$

\STATE Summarise $knowledge$ $graph$ $output$ using $summariser$ $model$ and set as $kgresultssummarised$

\STATE Chunk $knowledge$ $graph$ $output$ into $chunks$ of size maximum 128 tokens with 10 tokens overlap

\STATE Define $subanswer$ $context$ as empty string
\STATE Define $k$ as number of chunks to retrieve

\FOR{$chunk$ in $chunks$}
    \STATE Create a $sub-question$ from the $chunk$ using $Quantized LLM$ and $question$ $prompt$ 
    \STATE Retrieve $top$ $k$ $chunks$ from $BM25$ $store$ using $sub-question$
    \STATE Initialise $sub-context$ as empty string
    \FOR{$hybrid$ $chunk$ in $top$ $k$ $chunks$}
        \STATE Append $hybrid$ $chunk$ to $sub-context$
    \ENDFOR
    \STATE Generate $sub-answer$ using $Quantized LLM$ with $generation$ $prompt$ and passing in \textbf{user input} and $sub-context$
    \STATE Summarise $sub-answer$ using $summariser$ $model$ and set as $summarised$ $sub-answer$
    \STATE Append $summarised$ $sub-answer$ to $sub-answer$ $context$
\ENDFOR

\STATE Summarise $sub-answer$ $context$ using $summariser$ $model$ and set as $summarised-subanswer-context$

\STATE Set $full$ $context$ as concatenation of $kgresultssummarised$ and $summarised-subanswer-context$

\STATE Generate $answer$ using $Quantized LLM$ with $generation$ $prompt$ and passing in \textbf{user input} and $full$ $context$
\end{algorithmic}
\label{algo:k2ragprocess}
\end{algorithm}

\subsection{Implemented Naive RAG Pipelines}
We implemented 4 common naive RAG pipelines; 
\begin{itemize}
    \item \textbf{Semantic} - Retrieving only from a Dense Vector Database. 
    \item \textbf{Keyword} - Retrieving only from a Sparse Vector Database.
    \item \textbf{Hybrid} - Retrieving from both Dense and Sparse Vector Database. 
    \item \textbf{Knowledge Graph (KG)} - Retrieving only from a Knowledge Graph.
\end{itemize}

\subsubsection{Indexing } The indexing process for each information store is the same as in $K^2RAG$ but with chunks $C$ created with the unsummarized corpus $cor_{unsum}$ instead:

\begin{equation}
C = \bigcup_{d=1}^{|cor_{unsum}|} C_d \;|\; d\in cor_{unsum}
\label{formula:combineallunsum}
\end{equation}

\subsubsection{Retrieval and Generation } 
\begin{figure}[ht] \centering
\includegraphics[scale=0.4]{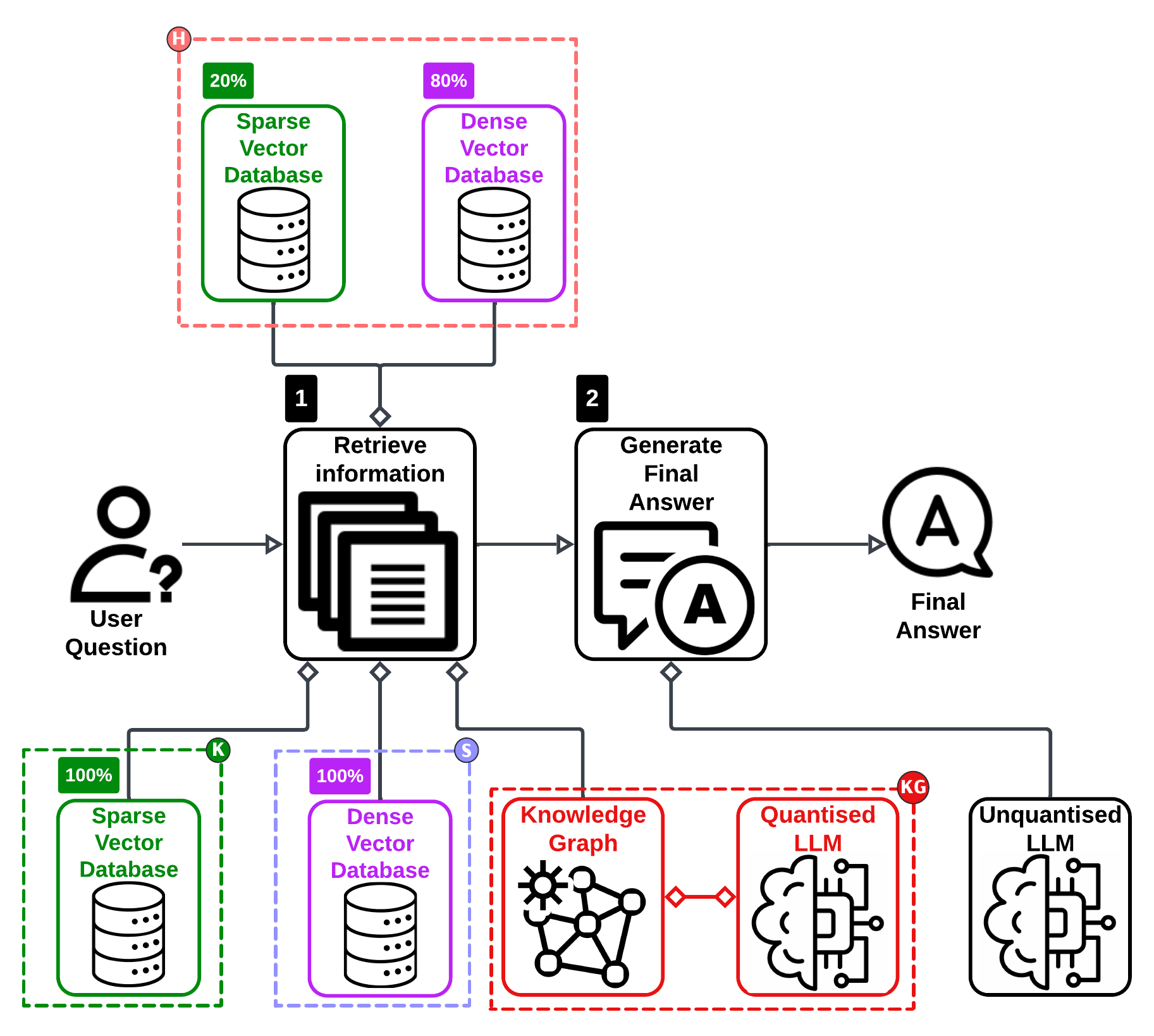}
\caption{Structure of competitor pipelines. Component (S): Semantic Search - Dense Vector Retrieval. Component (K): Keyword Search - Sparse Vector Retrieval. Component (H): Hybrid Search - Combination of Sparse and Dense Vector Retrieval. Component (KG): Knowledge Graph Retrieval}
\label{fig:competitorragretrieval} 
\end{figure}

All naïve retrieval-augmented pipelines in our framework follow a two-step process: (1) retrieval of information relevant to the input question, and (2) passage of the retrieved information to an unquantized large language model (LLM) for answer generation. In the case of our Semantic, Keyword, and Hybrid pipelines, we use a hybrid retrieval mechanism that combines semantic similarity and keyword matching. The weighting parameter $\lambda$ controls the balance between the two retrieval strategies: a value of $\lambda=1$ corresponds to purely semantic retrieval [Figure \ref{fig:competitorragretrieval}(S)], $\lambda=0$ corresponds to purely keyword-based retrieval [Figure \ref{fig:competitorragretrieval}(K)], and $\lambda=0.8$ gives higher preference to semantic retrieval while still incorporating keyword-based signals [Figure \ref{fig:competitorragretrieval}(H)].

For the naive Knowledge Graph (KG) retrieval pipeline, the retrieval process differs significantly. Instead of directly retrieving passages from the unsummarized corpus through vector search, we retrieve structured information from a knowledge graph that has been constructed using the unsummarized corpus and a quantized LLM. In this setup, the same quantized LLM is employed both for generating the knowledge graph and for querying it at retrieval time.

Across all pipelines, including the KG-based approach, the retrieved information is subsequently passed to an unquantized LLM for final answer generation. Importantly, all pipelines share a common answer generation prompt, which is the same as that used in $K^2RAG$ [Prompt \ref{prompt:answer_gen_prompt}]. This standardization of the answer generation step allows for a controlled comparison across different naive retrieval strategies with $K^2RAG$.

\subsection{Evaluation Framework}
\subsubsection{Dataset Selection and Description}
In order to perform a comprehensive evaluation on the proposed framework we have outlined two key requirements needed from the dataset as follows:
\begin{itemize}
    \item Dataset must contain question answer pairs in plain text format where answers are to be assumed as ground truth answers to the question.
    \item The question answer pairs must be constructed in reference to a document corpus which has all the required documents and their information from which ground truth answer to the questions are derived from.
\end{itemize}

Due to time and resource constraints we were unable to create our own dataset from scratch satisfying these requirements and hence the MultiHop-RAG dataset \cite{tang2024multihopragbenchmarkingretrievalaugmentedgeneration} was identified and used as it satisfies the above key requirements needed to perform the analysis.

The MultiHop-RAG dataset contains corpus data which was used for training along with a test dataset containing 2555 question answer pairs from which we performed evaluation of the pipeline. The corpus contains a 609 articles from a wide range of plain-text news articles data covering various topics such as entertainment, health, sports and technology \cite{tang2024multihopragbenchmarkingretrievalaugmentedgeneration} collected through the mediastack REST API and processed through a pipeline involving GPT-4 in order to review question answer pairs created from the articles by an LLM extracting factual statements from the articles and then GPT-4 generating these question answer pairs. This diversity in topics was also another reason this dataset was selected as it allows evaluating across multiple different domains of information to see if the pipeline can effectively answer questions across a wide variety of questions and their topics ensuring generalization.

\subsubsection{Evaluation Process}
\begin{figure}[H]
\centering
  \includegraphics[width=0.85\textwidth]{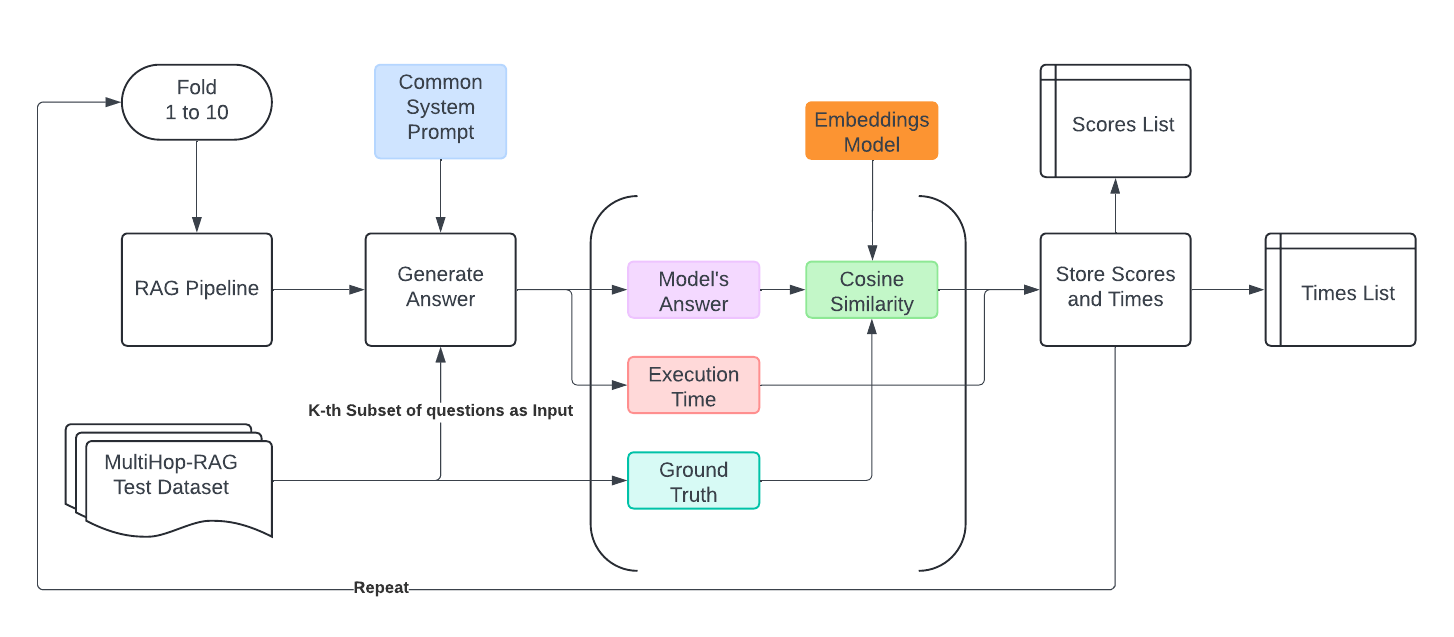}
  \caption{Evaluation Pipeline.}  
\label{fig:evaluationpipeline}
\end{figure}

We used a K-fold evaluation framework where we split the data into K=10 folds. For each pipeline, we generated the answers for each question in each fold and captured answering times for the questions to compare the execution times [Figure \ref{fig:evaluationpipeline}] [Algorithm \ref{algo:evaluationpipeline}]. 

To evaluate the accuracy of a pipeline's answer, we computed the similarity between the ground truth and the pipeline's output as a scoring function $s$ of the plaintext ground truth and pipeline output for each question defined as:

{\small
\begin{equation}
s(o,t) = \frac{e(o) \cdot e(t)}{|e(o)|*|e(t)|}
\label{formula:similarityscoring}
\end{equation}
}

where $o$ is the pipeline's output, $t$ is the ground truth and $e$ represents the embeddings model as a function to generate the vector embedding of a plaintext input.

We randomly partitioned the dataset into ten equal folds to simulate real-world usage of the pipeline. This approach reflects scenarios in which different users pose diverse questions to the system. Random assignment of test instances to folds helps mitigate the risk of performance bias, which could arise if certain folds contained clusters of similar questions that align particularly well with the strengths of the pipeline. For evaluation, we employed cosine similarity to assess the semantic similarity between the predicted and ground truth answers, as it effectively captures the degree of semantic similarity. Additionally, we recorded the execution time for each query to analyze the system's efficiency.

\begin{algorithm}
\caption{K-Fold Evaluation Algorithm}
\begin{algorithmic}[1]
\STATE Load in $rag$ $pipeline$
\STATE Load in $embeddings$ $model$
\STATE Load in $test$ $question-answer$ $pairs$
\STATE Initialize $scores$ as empty list
\STATE Shuffle test points in $test$ $question-answer$ $pairs$ randomly
\STATE Initialize $folds$ by assigning roughly equal number of points from $test$ $question-answer$ $pairs$ to 10 folds
\FOR{$fold$ $number$, $question, ground truth$ in $folds$}
    \STATE $start$ $time \leftarrow current$ $time$
    \STATE Generate $model$ $answer$ from $question$ using $rag$ $pipeline$
    \STATE $execution$ $time \leftarrow current$ $time$ - $start$ $time$
    \STATE Generate $ground$ $truth$ $vector$ using $embeddings$ $model$ by passing $ground$ $truth$
    \STATE Generate $model$ $answer$ $vector$ using $embeddings$ $model$ by passing $model$ $answer$
    \STATE Calculate $score$ by doing cosine similarity on $model$ $answer$ $vector$ and $ground$ $truth$ $vector$
    \STATE Append $fold$ $number$, $score$, $execution$ $time$ to $scores$ list
\ENDFOR
\end{algorithmic}
\label{algo:evaluationpipeline}
\end{algorithm}

\section{Results}\label{sec:results}
\subsection{Practical Implementation }
The Corpus Summarization and Evaluation processes were performed on a Linux machine with an NVIDIA L4 24GB VRAM GPU and 32GB RAM. We used the following tech stack for implementing all the pipelines; 

\begin{itemize}
    \item \textbf{Embeddings Model: } \textit{nomic-embed-text}
    \item \textbf{Quantized LLM: }\textit{Mistral-7B-Q4}
    \item \textbf{FP16 LLM: }\textit{Mistral-7B-FP16}
    \item \textbf{Summariser Model: }\href{https://huggingface.co/pszemraj/led-base-book-summary}{\textit{pszemraj/led-base-book-summary from HuggingFace}}
    \item \textbf{Sparse Vector Database: } \textit{Okapi BM25}
    \item \textbf{Dense Vector Database: } \textit{ChromaDB}
    \item \textbf{Knowledge Graph Implementation: } \textit{Microsoft GraphRAG} with \textit{Mistral-7B-Q4} as entity-relationship extractor and output generator.
\end{itemize}

\subsection{Corpus Summarization }
\begin{figure}[H]
\centering
  \includegraphics[width=0.55\textwidth]{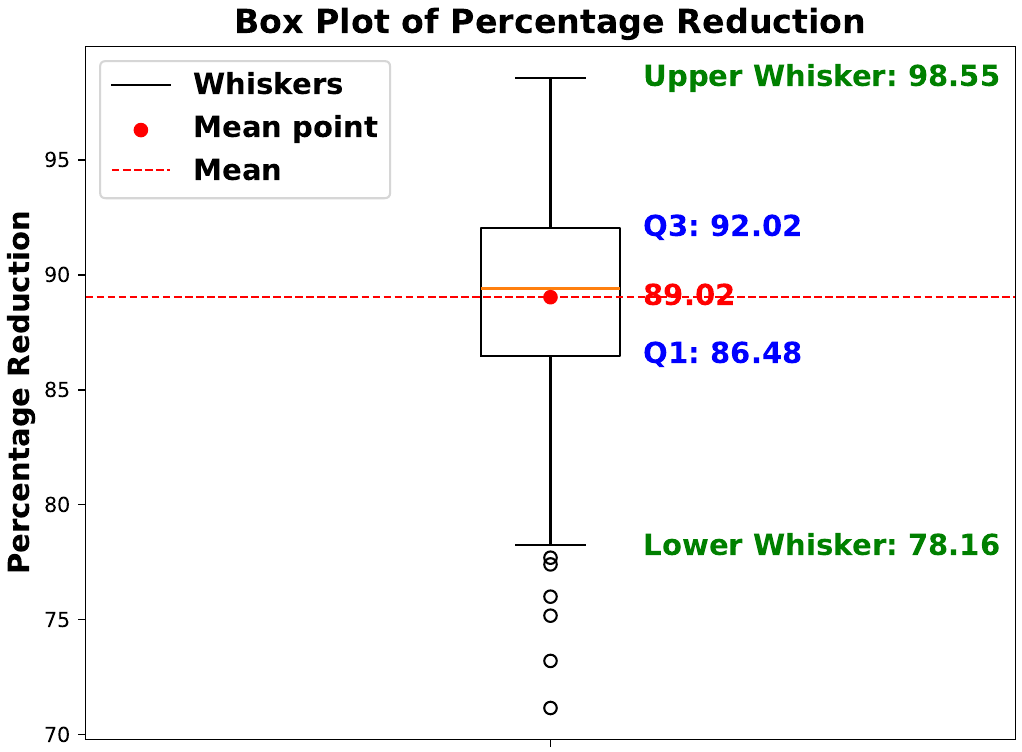}
  \caption{Boxplot of corpus size reduction.}  
\label{fig:boxplotsummarisationresults}
\end{figure}

Corpus summarization took roughly \textbf{25 minutes} to complete. Most documents have been reduced by roughly \textbf{86\%} to \textbf{92\%} while the overall mean reduction is roughly \textbf{89\%} [Figure \ref{fig:boxplotsummarisationresults}]. We attribute the consistent large reduction in size of the documents to considerable white space removed in many of the documents, leading to an effective reduction.

\subsection{Training Times }
\begin{figure}[H]
\centering
  \includegraphics[width=0.65\textwidth]{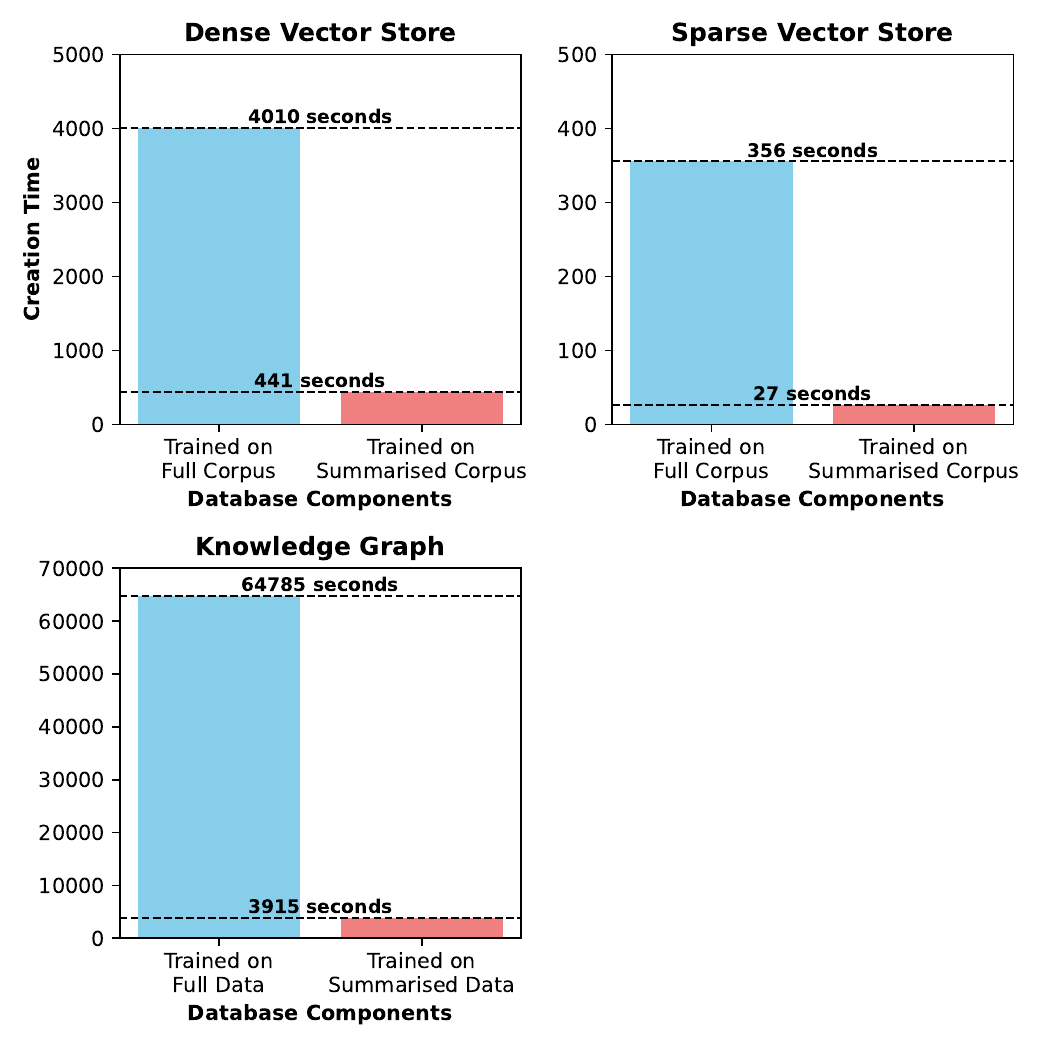}
  \caption{Information store component training times with summarized and unsummarized corpus. Training on the summarized corpus took significantly less time compared to unsummarized corpus.}
\label{fig:trainingtimes}
\end{figure}

By performing corpus summarization, sparse and dense vector stores, and knowledge graph creation times reduced by \textbf{89\%}, \textbf{97\%} and \textbf{94\%} respectively with significant time saved for knowledge graph creation from 18 hours to 1 hour [Figure \ref{fig:trainingtimes}] [Table \ref{table:trainingtimes}]. Hence, in exchange for \textbf{25 minutes} of performing summarization, training times were reduced on average by \textbf{93\%} across all information stores, successfully achieving Research Goal 1 in reducing creation times of information store components, particularly the knowledge graph. 

\begin{table}[H]
\centering
\begin{tabular}{lcl}
\toprule
\textbf{Information Store} & \textbf{Training Times} \\ \midrule
Dense Vector Full Corpus & 4010s  \\ \hline
Dense Vector Summarized Corpus & \textcolor{blue}{\underline{\textbf{441s}}} \\ \hline
Sparse Vector Full Corpus & 356s \\ \hline
Sparse Vector Summarized Corpus & \textcolor{blue}{\underline{\textbf{12s}}} \\ \hline
Knowledge Graph Full Corpus & 64785s \\ \hline
Knowledge Graph Summarized Corpus & \textcolor{blue}{\underline{\textbf{3915s}}} \\ \bottomrule
\end{tabular}
\caption{Table of Training Times}
\label{table:trainingtimes}
\end{table}

\vspace{-3em}

\subsection{Answer Accuracy }
\begin{figure}[H]
\centering
  \includegraphics[width=0.8\textwidth]{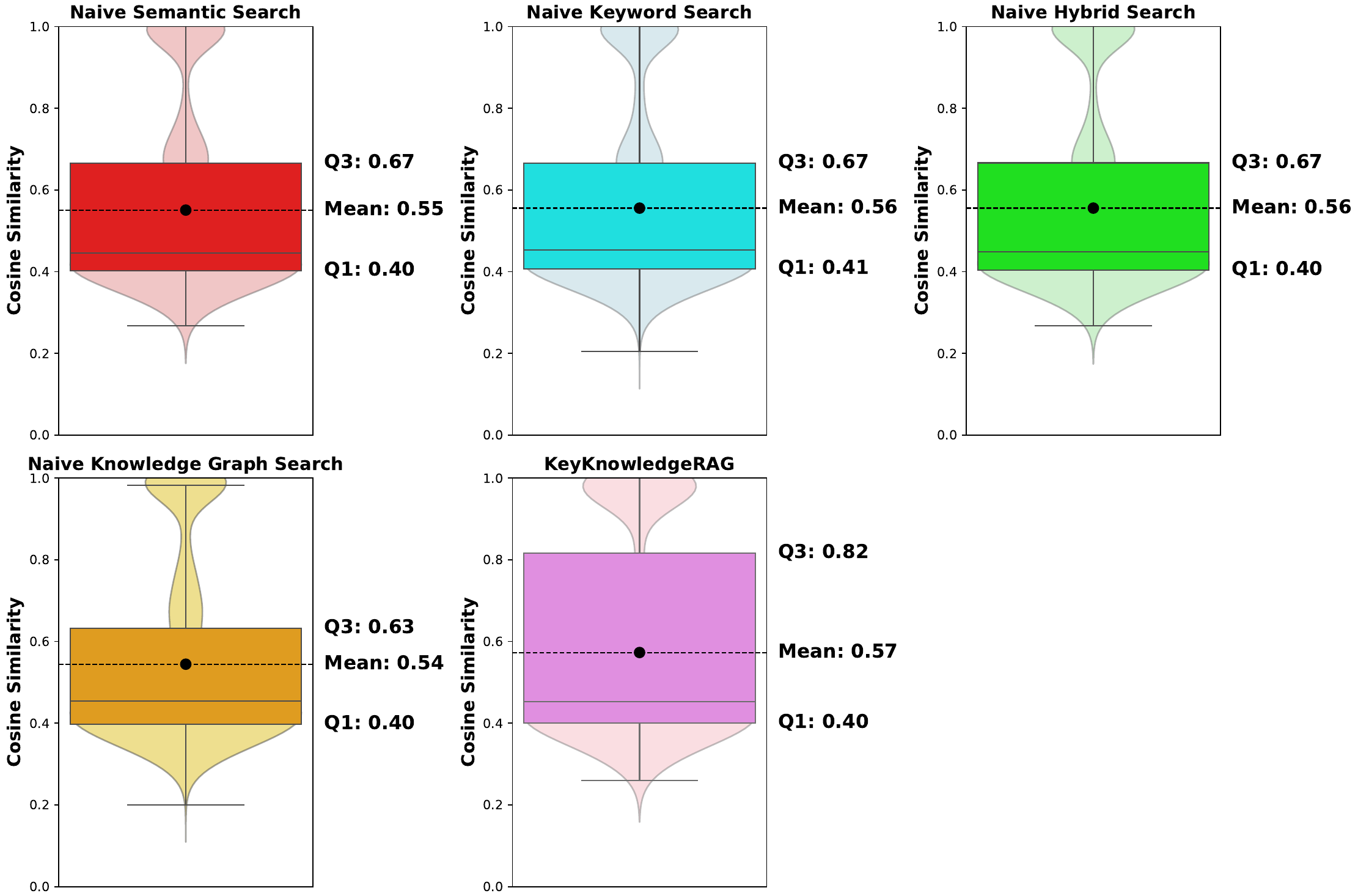}
  \caption{Pipeline Answer Accuracy results. $K^2RAG$ achieved the best performance with highest mean and 3rd Quartile value.}
\label{fig:answersimilarity}
\end{figure}

We observe that the score distributions for the Semantic, Keyword, Hybrid Search, and KG pipelines are very similar [Figure \ref{fig:answersimilarity}]. This suggests that although these naive pipelines utilize different components, they cannot fully exploit their potential for Question Answering, which is why they are considered naive. In contrast, the distribution for our $K^2RAG$ framework demonstrates a significantly larger $Q_3$ quartile compared to the naive frameworks, which indicates that $K^2RAG$ was able to answer more questions with greater similarity to the ground truths. $K^2RAG$ has a slightly higher mean accuracy of \textbf{0.57} compared to other naive methods despite a high $Q_3$ quartile value [Table \ref{table:accuracies}]. This implies that $K^2RAG$ is better than naive RAG pipelines in answering certain types of questions which pushed the $Q_3$ quartile value up, but there were not enough of these types of questions to induce a greater increase in the mean similarity.
Overall, $K^2RAG$ achieved the highest mean similarity score of \textbf{0.57} and the top $Q_3$ quartile score of \textbf{0.82} [Table \ref{table:accuracies}], successfully achieving Research Goals 2 and 3 by retrieving only the most relevant information and creating better contexts for generating accurate and information-rich answers.

\vspace{-1em}
\begin{table}[H]
\centering
\begin{tabular}{lcll}
\toprule
\textbf{Pipeline} & \textit{\textbf{$Q_1$}} & \textbf{Mean} & \textit{\textbf{$Q_3$}} \\ \midrule
Naive Semantic Search & 0.40 & 0.55 & 0.67 \\ \hline
Naive Keyword Search & \underline{\textbf{0.41}} & 0.55 & 0.67 \\ \hline
Naive Hybrid Search & 0.4 & 0.56 & 0.67 \\ \hline
Naive Knowledge Graph Search & 0.4 & 0.54 & 0.4 \\ \hline
\textcolor{blue}{$K^2RAG$} & \textcolor{blue}{0.4} & \textcolor{blue}{\underline{\textbf{0.57}}} & \textcolor{blue}{\underline{\textbf{0.82}}} \\ \bottomrule
\end{tabular}
\caption{Table of Pipeline Answer $K^2RAG$ achieved the best performance with highest mean and 3rd Quartile value.}
\label{table:accuracies}
\end{table}

\subsection{Execution Times }
\begin{figure}[H]
\centering
  \includegraphics[width=0.8\textwidth]{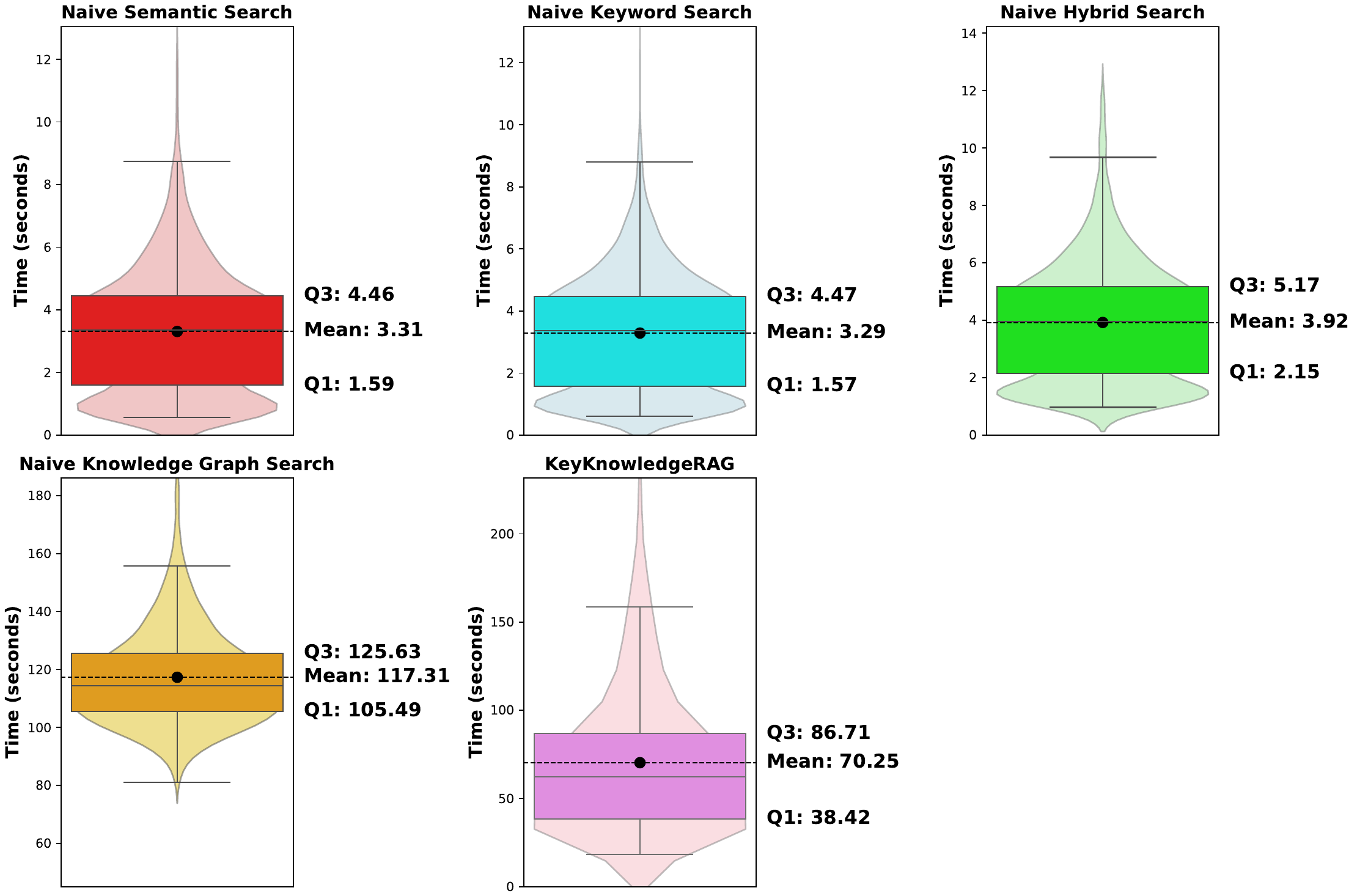}
  \caption{Pipeline Execution Times results. $K^2RAG$ is faster than KG Search despite more steps.}
\label{fig:executiontimes}
\end{figure}

Semantic Search, Keyword Search, and Hybrid Search were the fastest running pipelines. KG Search recorded the longest mean runtime at 117.31 seconds, with a small interquartile range, indicating consistent performance [Figure \ref{fig:executiontimes}]. The longer runtime can be attributed to the large, unsummarized knowledge graph used in retrieval. Our proposed $K^2RAG$ framework achieved a mean runtime of 70.25 seconds, significantly faster than naive KG Search but slower than vector based search methods such as semantic, keyword and hybrid search [Table \ref{table:executiontimes}]. This is expected as querying from the knowledge graph component takes more time as it is not just a simple retrieval based on matching vectors. $K^2RAG$ exhibited a wider inter-quartile range due to the sub-questions step, which increases the number of chunks as the size of the knowledge graph results grows. Despite having more steps, $K^2RAG$ had faster execution times than KG Search and hence successfully achieving Research Goal 4 with faster responses. 

\begin{table}[H]
\centering
\begin{tabular}{lcl}
\toprule
\textbf{Pipeline} & \textbf{Mean Execution Time}\\ \midrule
Naive Semantic Search & 3.31s  \\ \hline
Naive Keyword Search & \underline{\textbf{3.29s}} \\ \hline
Naive Hybrid Search & 3.92s \\ \hline
Naive Knowledge Graph Search & 117.31s \\ \hline
\textcolor{blue}{$K^2RAG$} & \textcolor{blue}{70.25s} \\ \bottomrule
\end{tabular}
\caption{Table of Execution Times. $K^2RAG$ is faster than KG Search despite more steps.}
\label{table:executiontimes}
\end{table}

\subsection{Memory Footprint }
The memory footprint analysis, as presented in Table \ref{table:memoryfootprint}, highlights the significant reduction in VRAM consumption achieved by $K^2RAG$ compared to other retrieval pipelines. Specifically, while the Semantic Search Pipeline, Keyword Search Pipeline, and Hybrid Search Pipeline each require 14.3GB of memory, and the Knowledge Graph Search demands an even higher 18.1GB. Meanwhile $K^2RAG$ operates with only 5GB of VRAM [Table \ref{table:memoryfootprint}].

This reduction represents a mean threefold decrease in memory usage relative to the other pipelines, demonstrating a substantial optimization in resource efficiency. Such an improvement directly aligns with Research Goal 4, which emphasizes minimizing computational resource requirements.

\begin{table}[H]
\centering
\begin{tabular}{lcl}
\toprule
\textbf{Pipeline} & \textbf{Memory Footprint}\\ \midrule
Semantic Search Pipeline & 14.3GB  \\ \hline
Keyword Search Pipeline & 14.3GB \\ \hline
Hybrid Search Pipeline & 14.3GB \\ \hline
Knowledge Graph Search & 18.1GB \\ \hline
\textcolor{blue}{KeyKnowledgeRAG} & \underline{\textcolor{blue}{\textbf{5GB}}} \\ \bottomrule
\end{tabular}
\caption{Table of Memory Footprints}
\label{table:memoryfootprint}
\end{table}

\vspace{-2em}

\section{Discussion} \label{sec:discussion}
\subsection{Achieving Research Goal 1: Reduce information store creation and update times}

$K^2RAG$ effectively reduces the creation times for Knowledge Graph and Sparse and Dense vector databases, such as the Semantic and BM25-based Keyword databases. Performing text summarization on the training corpus led to an average reduction in training times of up to \textbf{93\%}, with only \textbf{25 minutes} spent on summarization.

For instance, the Semantic database creation, which originally took \textbf{4010 seconds (67 minutes)}, was reduced to \textbf{441 seconds (7.35 minutes)} when using the summarized corpus. Even when accounting for the summarization process, the total time was \textbf{32.25 minutes}, significantly lower than training with the full corpus. Similarly, Knowledge Graph creation time dropped from nearly \textbf{18 hours} to around \textbf{1 hour}, thanks to the reuse of the summarized corpus. This approach not only saves time and resources but also reduces the cost of LLM-based Knowledge Graph creation, which is typically expensive to host.

\subsection{Achieving Research Goal 2: Mitigate the Needle-in-a-Haystack problem}

$K^2RAG$ effectively mitigates the "Needle in a Haystack" problem, which can hinder LLM performance when handling large contexts. By integrating the Longformer-based \textit{led-base-book-summary} summarizer at key stages of the retrieval pipeline, the framework reduces context size while preserving essential information.

$K^2RAG$ maintained answer quality and even outperformed naive pipelines on certain questions, as reflected in the higher accuracy and $Q_3$ quantile. This approach allows for improved LLM performance without a bloated context, preventing the model from missing critical details and enhancing the final answer quality.

\subsection{Achieving Research Goal 3: Improve retrieval accuracy}

The $K^2RAG$ framework enhances passage retrieval through two key components. First, the knowledge component identifies relevant topics for answering the question. Based on these results, the system further explores each topic within the knowledge graph to extract more concrete and relevant chunks. This approach enables the development of more comprehensive answers. Similar to Research Goal 2, this strategy improves the quality of the final LLM-generated responses.

\subsection{Achieving Research Goal 4: Lower Temporal and Computational Overhead}

While rerankers enhance passage retrieval accuracy, they require an additional large model, increasing resource demands. Similarly, the naive Knowledge Graph RAG pipeline depends on two LLMs: \textit{Mistral-7B-FP16} (14GB VRAM) and \textit{Mistral-7B-Q4} (4.1GB VRAM).

In contrast, the $K^2RAG$ framework streamlines the process by utilizing only \textit{Mistral-7B-Q4} (4.1GB VRAM) for querying the Knowledge Graph, question generation, and final answer generation, along with a small summarization model requiring just 648MB VRAM. This results in a total VRAM usage of 4.8GB, allowing for efficient RAG pipelines on smaller machines while significantly reducing computational and monetary costs.

Moreover, $K^2RAG$ achieves faster response times and superior answer quality compared to naive RAG pipelines. This is evident in its reduced question-answering times relative to the naive Knowledge Graph Search RAG, enabling $K^2RAG$ to process more queries in a span of time.

\subsection{Limitations} \label{sec:limitation}
While our RAG framework shows a slight mean accuracy advantage over naive methods, a mean similarity of 0.57, despite a high \(Q_3\) value, indicates limited performance gains on most questions. This suggests it excels only at certain question types, but the narrow scope restricts broader improvement. Reliance on broad knowledge graph searches may also reduce precision. A more focused retrieval strategy could yield more precise topic pinpointing and higher accuracy.

\section{Conclusion} \label{sec:conclusion}
This paper introduces the $K^2RAG$ framework, which addresses key limitations of traditional RAG methods. By utilizing a summarized corpus, we achieved an average reduction of 93\% in training times. When combined with quantized models and an advanced retrieval process, $K^2RAG$ outperformed all tested naive RAG methods in terms of answer accuracy, resource efficiency, and execution speed, even with more complex steps. As a result, $K^2RAG$ offers a more accurate and lightweight RAG pipeline. While the framework demonstrated a slight improvement in mean similarity and a higher $Q_3$ value, it primarily excelled in answering certain question types. For future work, we recommend testing $K^2RAG$ on additional datasets to assess its performance on a wider range of topics, and enhancing knowledge graph retrieval techniques to further improve answer accuracy across all question types by identifying a broader set of relevant topics.

\bibliographystyle{unsrt}  
\bibliography{references}

\end{document}